\documentclass[fleqn,10pt]{wlpeerj}
\usepackage{amssymb}
\usepackage{latexsym}

\usepackage{amsmath}
\usepackage{bm}
\usepackage{threeparttable}
\usepackage{algorithm}
\usepackage{algpseudocode}

\algblockdefx[NAME]{Block}{Endblock}[2][]{\{{\bf #2}\}}{}

\newcommand{\argmin}{\mathop{\rm arg~min}\limits}

\title{Approximate spectral clustering using both reference vectors and topology of the network generated by growing neural gas}
\author[1,2]{Kazuhisa Fujita}

\affil[1]{Komatsu University, 10-10 Doihara-Machi, Komatsu, Ishikawa, Japan 923-0921}
\affil[2]{University of Electro-Communications, 1-2-1 Chofu-gaoka, Chofu, Tokyo, Japan 182-8585}

\corrauthor[1]{Kazuhisa Fujita}{kazu@spikingneuron.net}

\begin{abstract}
  Spectral clustering (SC) is one of the most popular clustering methods and often outperforms traditional clustering methods.
  SC uses the eigenvectors of a Laplacian matrix calculated from a similarity matrix of a dataset.
  SC has serious drawbacks: the significant increases in the time complexity derived from the computation of eigenvectors and the memory space complexity to store the similarity matrix.
  To address the issues, I develop a new approximate spectral clustering using the network generated by growing neural gas (GNG), called ASC with GNG in this study.
  ASC with GNG uses not only reference vectors for vector quantization but also the topology of the network for extraction of the topological relationship between data points in a dataset.
  ASC with GNG calculates the similarity matrix from both the reference vectors and the topology of the network generated by GNG.
  Using the network generated from a dataset by GNG, ASC with GNG achieves to reduce the computational and space complexities and improve clustering quality.
  In this study, I demonstrate that ASC with GNG effectively reduces the computational time.
  Moreover, this study shows that ASC with GNG provides equal to or better clustering performance than SC.
\end{abstract}

\begin{document}

\flushbottom
\maketitle
\thispagestyle{empty}

%% main text
\section{Introduction}
\label{Introduction}

A clustering method is a workhorse and becomes more important for data analysis, data mining, image segmentation, and pattern recognition.
The most famous clustering method is \textit{k}-means, but it cannot accurately partition a nonlinearly separable dataset.
Spectral clustering (SC) is one of the efficient clustering methods for a nonlinearly separable dataset \citep{Filippone:2008} and can extract even complex structures such as half-moons data \citep{Bojchevski:2017}.
SC often outperforms traditional popular clustering methods such as \textit{k}-means \citep{vonLuxburg:2007}.
However, SC has significant drawbacks: the considerable increases in computational complexity and space complexity with the number of data points.
These drawbacks make it difficult to use SC for a large dataset.
Nowadays, the drawbacks are becoming more crucial as datasets become more massive, more varied, and more multidimensional.

SC treats a dataset as a graph (network) consisting of nodes and weighted edges.
The nodes and the edges respectively represent the data points in the dataset and the connections between the data points.
The weights of the edges are the similarities between data points.
In SC, eigenvectors are calculated from the Laplacian matrix derived from the similarity matrix of the network.
The rows of the matrix that consist of the eigenvectors are clustered using a traditional clustering method such as \textit{k}-means.
The drawbacks of SC for a large dataset are the time complexity to compute the eigenvectors and the space complexity to store the similarity matrix.

Researchers have tackled the high computational cost of SC.
There are four approaches to improve the computational cost.
The first approach is parallel computing to reduce computational time \citep{Song:2008,Chen:2011,Jin2013}.
The second approach is data size reduction by random sampling.
For example, Sakai and Imiya \citep{Sakai:2009} make a similarity matrix small by random sampling its columns to reduce the computational cost of eigendecomposition.
The third approach is to use a low-rank matrix that approximates the similarity matrix of the original dataset to avoid calculating the whole similarity matrix \citep{Fowlkes:2004,Li:2011}.
The last approach is to reduce a data size using a vector quantization method such as \textit{k}-means \citep{Yan:2009}, self-organizing map (SOM) \citep{Duan:2012} and neural gas (NG) \citep{Moazzen:2016}.
This method is called approximate spectral clustering or two-level approach.
In approximate spectral clustering (ASC), data points are replaced with fewer reference vectors.
We can decrease the computational cost of SC by reducing data size using a quantization method.

This study focuses on ASC using SOM and its alternatives such as NG and growing neural gas (GNG).
SOM and its alternatives are brain-inspired artificial neural networks that represent data points in the network parameters.
The network consists of units that have weights regarded as reference vectors and edges connecting pairs of the units.
ASC uses fewer reference vectors instead of data points to reduce the size of the input to SC.
Furthermore, the reference vectors are regarded as local averages of data points, thus, less sensitive to noise than the original data \citep{Vesanto:2000}.
This may improve the clustering performance of SC.
The performance of SC highly depends on the quality of the constructed similarity matrix from input data points \citep{Chang:2008,Li:2018,Park:2018,Zhang:2011}. 
In other words, to construct a more robust similarity matrix is to improve the clustering performance of SC \citep{Lu:2016}.
Because of the dependence on the quality of the similarity matrix, SC is highly sensitive to noisy input data \citep{Bojchevski:2017}.
Thus, the lower sensitivity of reference vectors to noise may improve the clustering quality of SC.
In many studies, reference vectors of a network generated by SOM are used to reduce input size, but the topology of the network is not exploited.

In this study, I develop a new ASC using a similarity matrix calculated from both the reference vectors and the topology of the network generated by GNG, called ASC with GNG.
The key point is to regard the network generated by GNG as a similarity graph and calculate the similarity matrix from not only the reference vectors but also the topology of the network that reflects the topology of input.
I employ GNG to generate a network because the network generated by GNG can represent important topological relationships in a given dataset \citep{Fritzke:1995} and the better similarity matrix will lead to better clustering performance in the ASC.
The quantization by GNG reduces the computational complexity and the space complexity of SC.
Furthermore, the effective extraction of the topology by GNG may improve clustering performance.
This paper investigates the computation time and the clustering performance of ASC with GNG.
Moreover, I compare ASC with GNG with ASC using a similarity matrix calculated from quantization results generated by neural gas, Kohonen's SOM, and \textit{k}-means instead of GNG.

\section{Related works}
\label{sec:related_works}

The most well-known clustering method is \textit{k}-means \citep{MacQueen:1967}, which is one of the top 10 most common algorithms used in data mining \citep{Wu:2007}.
\textit{k}-means is highly popular because it is simple to implement and yet effective in performance \citep{Haykin:2009}.
It groups data points so that the sum of Euclidean distances between the data points and their centroids is small.
It can accurately partition only linearly separable data.

Expectation–maximization algorithm (EM algorithm) is a popular tool for simplifying difficult maximum likelihood problems \citep{Hastie:2009}.
EM algorithm can estimate the parameters of a statistical model and is also used for clustering.
When we use EM algorithm for clustering, we often assume that data points are generated from a mixture of Gaussian distributions.
However, EM algorithm cannot precisely partition data points not generated from the assumed model.
\textit{k}-means is described by slightly extending the mathematics of the EM algorithm to this hard threshold case \citep{Bottou:1994} and assumes a mixture of isotropic Gaussian distributions with the same variances.

A major drawback to k-means is that it cannot partition nonlinearly separable clusters in input space \citep{Dhillon:2004}.
There are two approaches for achieving nonlinear separations using \textit{k}-means.
One is kernel \textit{k}-means \citep{Girolami:2002} that partitions the data points in a higher-dimensional feature space after the data points are mapped to the feature space using the nonlinear function \citep{Dhillon:2004}.
All the computation of kernel \textit{k}-means can be done by the kernel function using the kernel trick \citep{Ning:2016}.
However, kernel \textit{k}-means cannot accurately partition data points if the kernel function is not suitable for the data points.
The other is \textit{k}-means using other distances such as spherical \textit{k}-means (sk-means) \citep{Dhillon:2001,Banerjee:2003,Banerjee:2005} and cylindrical \textit{k}-means (cyk-means) \citep{Fujita:2017}.
Sk-means uses cosine distance because of assuming the data points generated from a mixture of von Mises-Fisher distributions.
Cyk-means uses the distance that combines cosine distance and Euclidean distance because of assuming the data points generated from a mixture of joint distributions of von Mises distribution and Gaussian distribution. 
However, $k$-means using other distances cannot accurately partition data points if the distance is not suitable for the data points.

Affinity propagation \citep{Frey2007} is a clustering method using a similarity matrix as in SC and is derived as an instance of the max-sum algorithm.
Affinity propagation simultaneously considers all data points as potential exemplars and recursively transmits real-value messages along edges of the network until a good set of exemplars is generated \citep{Frey2007}.
Its advantage is that there is no need to pre-specify the number of clusters and no assumption of a mixture of distributions.
Its time complexity is $O(N^2T)$ \citep{Fujiwara:2011,Khan:2018}, where $N$ and $T$ are the number of data points and the number of iterations, respectively.
Space complexity is $O(N^2)$ because it stores a similarity matrix.
For a large dataset, these complexities are not ignored.

Spectral clustering (SC) is a popular modern clustering method based on eigendecomposition of a Laplacian matrix calculated from a similarity matrix of a dataset \citep{Tasdemir:2015}.
SC does not assume a statistical distribution and partitions a dataset using only a similarity matrix.
SC displays high performance for clustering nonlinear separable data \citep{Chin:2015} and has been applied to various fields such as image segmentation \citep{Eichel:2013}, co-segmentation of 3D shapes \citep{Luo:2013}, video summarization \citep{Cirne:2013}, identification of cancer types \citep{Chin:2015,Shi:2017}, document retrieval \citep{Szymanski:2017}.
In SC, a dataset is converted to a Laplacian matrix calculated from the similarity matrix of the dataset.
We obtain a clustering result by grouping the rows of the matrix that consists of the eigenvectors of the Laplacian matrix.
SC often outperforms a traditional clustering method, but it requires enormous computational cost and large memory space for a large dataset.
Especially, its use is limited since it is often infeasible due to the computational complexity of $O(N^3)$ \citep{Izquierdo-Verdiguier:2015,Tasdemir:2012,Tasdemir:2015,Wang:2013}, where $N$ is the number of data points.
The huge computational complexity of SC is mainly derived from the eigendecomposition and constitutes the real bottleneck of SC for a large dataset \citep{Izquierdo-Verdiguier:2015}.
The required memory space increases with $O(N^2)$ \citep{Mall:2013} because the similarity matrix is an $N \times N$ matrix.

Clustering a large dataset often becomes more challenging due to increasing computational cost with the size of a dataset.
One approach to reducing the computational cost of clustering is two-level approach that partitions the quantization result of a dataset \citep{Vesanto:2000}.
In two-level approach, first, data points are converted to fewer reference vectors by a quantization method such as SOM and \textit{k}-means (abstraction level 1).
Then the reference vectors are combined to form the actual clusters (abstraction level 2).
Each data point belongs to the same cluster as its nearest reference vector.
The two-level approach has the advantage of dealing with fewer reference vectors instead of the data points as a whole, therefore reducing the computational cost \citep{Silva:2014}.
This approach does not limit a quantization method in abstraction level 1 and a clustering method in abstraction level 2.
There are many pairs of quantization methods and clustering methods, for example, SOM and hierarchical clustering \citep{Vesanto:2000,Tasdemir:2011}, GNG and hierarchical clustering \citep{Mitsyn:2011}, and SOM and normalized cut \citep{Yu:2014}.
Especially, the approach using SC in abstraction level 2 is called Approximate Spectral Clustering (ASC) \citep{Tasdemir:2015}.

In two-level approach, there is the concern of underperformance of clustering by the quantization because it uses fewer reference vectors instead of all the data points in a dataset.
However, a quantized dataset would be sufficient in many cases \citep{Bartkowiak:2005}.
Furthermore, two-level approach using SOM also has the benefit of noise reduction \citep{Vesanto:2000}.

\section{Approximate spectral clustering with growing neural gas} % (fold)

% l
% M: the number of units
% n: neighbor unit
% N: data size
% i: unit number
% j: unit number
% k: the number of clusters
% x: data point
% d: dimension
% A: similarity matrix
% a: element of A
% L: Laplacian matrix
% D: diagonal matrix
% s1: winning neuron
% s2: second winner

This paper proposes approximate spectral clustering with growing neural gas (ASC with GNG).
ASC with GNG partitions a dataset using a similarity matrix calculated from both reference vectors and the topology of the network generated by GNG.
ASC with GNG consists of the two processes, which are abstraction level 1 and abastraction level2, based on two-level approach \citep{Tasdemir:2015,Vesanto:2000}.
In abstraction level 1, a dataset is converted into the network by GNG.
The similarity matrix is calculated from the network, considering the reference vectors and the topology of the network.
In abstraction level 2, the reference vectors are merged by SC using the similarity matrix.
The data points in the dataset are assigned to the clusters to which the reference vectors nearest to the data points belong.

ASC with GNG can partition a nonlinearly separable dataset.
After abstraction level 2, the data points are assigned to the nearest reference vectors based on Euclidean distance.
Thus, ASC with GNG divides the space into regions consisting of Voronoi polygons around reference vectors.
In ASC with GNG, SC merges the reference vectors to $k$ clusters.
Simultaneously, the Voronoi polygons corresponding to the reference vectors are also merged to $k$ regions.
ASC with GNG provides the complex decision regions consisting of Voronoi polygons and achieves nonlinear separation.

At first, ASC with GNG appears to be the same as other ASCs because many ASCs using a quantization method such as \textit{k}-means, SOM and its alternatives already exist.
For example, \cite{Duan:2012} have proposed ASC using SOM.
In their method, data points are quantized by SOM, and SC partitions the reference vectors.
The similarity measure of their method is defined by the Euclidean distance divided by local variance reflecting the distribution of data points around a reference vector.
The interesting point of their method is using a local variance.
\cite{Moazzen:2016} have proposed ASC using NG.
Similarly, in their method, the similarity matrix used by SC is calculated from reference vectors.
It is interesting that their method uses integrated similarity criteria to improve accuracy.
In ASC with GNG as well as the other ASCs, reference vectors are partitioned by SC.
However, ASC with GNG uses both reference vectors and topology of the network generated by GNG, but the other ASCs use only reference vectors.
Using the topology will play an important role in improving clustering performance because SC can be regarded as a type of partition problem for a network \citep{Diao:2015} and partitions the units of the network into disjoint subsets.

\subsection{Algorithm}
\label{sub:algorithm}

ASC with GNG consists of generating a network by GNG, partitioning reference vectors by SC, and assigning data points into clusters.
The algorithm for ASC with GNG is given in Algorithm \ref{alg:ASC}.

\begin{figure}[tb]
  \begin{center}
    \includegraphics[width=140mm]{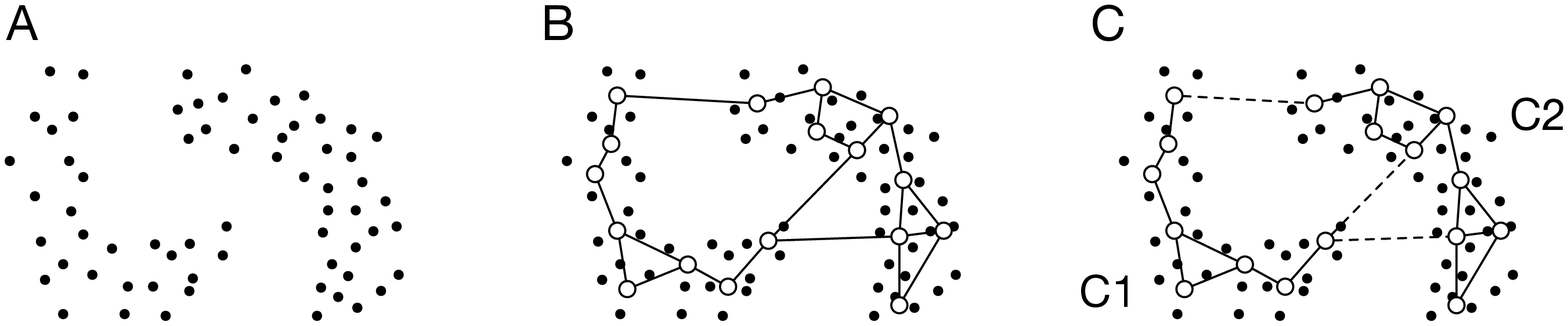}
    \caption{Schematic image of ASC with GNG.
    A. Data distribution. The dots denote the data points.
    B. Generated network. The open circles and the solid lines denote the units and the edges, respectively.
    C. Partition of the units using SC. The broken line denotes the connection cut by SC.
    SC merges the units into two clusters: C1 and C2.
    In other words, SC removes the edges connecting the units that belong to different clusters.}
    \label{fig:scheme}
  \end{center}
\end{figure}

Let us consider a set of $N$ data points, $X = \{\bm x_1, \bm x_2, ..., \bm x_n, ..., \bm x_N\}$, where $\bm x_n \in R^d$ (Fig. \ref{fig:scheme}A).
The data point $n$ is denoted by $\bm x_n$.
In this study, each data point is previously normalized by $\bm x_n \leftarrow \bm x_n / \| \bm x_\mathrm{max} \|$, where $\bm x_\mathrm{max}$ is the data point that has the maximum norm.
For an actual application, such normalization will not be required because we can use specific parameters for the application.

GNG generates a network from data points (Fig. \ref{fig:scheme}B).
The generated network is not a complete graph, and its topology reflects input space.
The unit $i$ in the network has the reference vector $\bm{w}_i \in R^d$.
The algorithm of GNG is denoted in Appendix \ref{sec:gng}.

The similarity matrix $A \in R^{M \times M}$ is calculated from the reference vectors and the topology of the generated network.
$M$ is the number of units in the network.
If the unit $i$ connects to the unit $j$, the element $a_{ij}$ of $A$ is defined as follows:
\begin{equation}
\label{eq:gaussian}
  a_{ij} = \exp \Big(- \frac{\|\bm{w}_i - \bm{w}_j \|^2}{2\sigma^2}\Big),
\end{equation}
where $\bm{w}_i$ and $\bm{w}_j$ are the reference vectors of $i$ and $j$, respectively.
Otherwise, $a_{ij} = 0$.
Equation \ref{eq:gaussian} is the Gaussian similarity function that is most widely used to obtain a similarity matrix.
The proposed method's new point is to make a similarity matrix from both the reference vectors and the network topology.
The similarity matrix is sparse because of using the network topology.
However, the similarity matrix has information about significant connections representing input space.

The reference vectors are merged to $k$ clusters by SC as shown in Fig. \ref{fig:scheme}C.
The normalized Laplacian matrix $L_\mathrm{sym} \in R^{M \times M}$ is calculated from $A$.
Here, we define the diagonal matrix $D \in R^{M \times M}$ to calculate the Laplacian matrix.
The element of the diagonal matrix is $d_i = \sum_{j = 1}^M a_{ij}$.
The Laplacian matrix is $L = D - A$.
We derive the normalized Laplacian matrix from the following equation:
\begin{equation}
\label{eq:normLaplacian}
  L_\mathrm{sym} = D^{-1/2}LD^{-1/2}.
\end{equation}
Here, we calculate the $k$ first eigenvectors $\bm u_1, ..., \bm u_k$ of $L_\mathrm{sym}$.
Let $U \in R^{M\times k}$ be the matrix containing the vector $\bm u_1, ..., \bm u_k$ as columns.
For $i = 1, ..., M$, let $\bm y_i \in R^k$ be the vector corresponding to the $i$-th row of $U$.
\textit{k}-means assigns $(\bm y_i)_{i=1, ..., M}$ to clusters $C_1, ..., C_k$.
Each row vector corresponds with each reference vector one-to-one.

Each data point is assigned to the cluster to which the nearest reference vector belongs.
Finally, the data points are assigned to the clusters.

\begin{algorithm}
\caption{ASC with GNG}
\label{alg:ASC}
\begin{algorithmic}[1]
\Require Data points $X \in R^d$, the number of clusters $k$
\Ensure A partition of given data points into $k$ clusters $C_1, ..., C_k$
\State \{Abstraction Level1\}
\State \hspace{10pt} Generate the network from the dataset $X$ using GNG.
\State \hspace{10pt} Calculate the similarity matrix $A \in R^{M \times M}$ from the reference vectors and the topology of the network.
\State \{Abstraction Level 2\}
\State \hspace{10pt} Calculate the normalized Laplacian matrix $L_\mathrm{sym}$ by Eq. \ref{eq:normLaplacian}.
\State \hspace{10pt} Calculate the $k$ first eigenvectors $\bm u_1, ..., \bm u_k$ of $L_\mathrm{sym}$.
\State \hspace{10pt} Let $U \in R^{M\times k}$ be the matrix containing the vectors $\bm u_1, ..., \bm u_k$ as columns.
\State \hspace{10pt} For $i = 1, ..., M$, let $\bm y_i \in R^k$ be the vector corresponding to the $i$-th row of $U$.
\State \hspace{10pt} Assign $(\bm y_i)_{i=1, ..., M}$ to clusters $C_1, ..., C_k$ by \textit{k}-means.
\State \{Assign data points to the clusters.\}
\State \hspace{10pt} Find the nearest unit to each data point.
\State \hspace{10pt} Each data point is assigned to the cluster to which the nearest unit is assigned.
\end{algorithmic}
\end{algorithm}

\subsection{ASC with SOM and its alternatives}

In this study, ASC with GNG is compared with ASCs with NG, SOM, \textit{k}-means, and GNG using no topology to investigate the difference in clustering performances depending on the methods used to generate the network.
ASCs with NG and SOM use NG and Kohonen's SOM to generate networks instead of GNG, respectively.
The weights of the network generated by NG and SOM are calculated using the Gaussian similarity function.
ASC with \textit{k}-means uses centroids generated by \textit{k}-means as units in the network.
The network is fully connected, and its weights are calculated using the Gaussian similarity function.
ASC with GNG using no topology uses only the reference vectors generated by GNG.
In ASC with GNG using no topology, the network is fully connected, and its weights are calculated using the Gaussian similarity function.
In this study, ASC with GNG is also compared with SC.
SC uses a fully-connected network, the Gaussian similarity function, a normalized Laplacian matrix $L_\mathrm{sym} = D^{-1/2}LD^{-1/2}$, and \textit{k}-means.

\subsection{Complexity} % (fold)
\label{sub:complexity}

The time complexity of SC is $O(N^3)$ \citep{Izquierdo-Verdiguier:2015,Tasdemir:2015,Wang:2013}, where $N$ is the number of data points.
The complexity relies on the eigendecomposition of a Laplacian matrix.
The eigendecomposition takes $O(N^3)$ time.

The time complexity of ASCs with GNG is $O(MT + M^3 + NM)$, where $M$ and $T$ are respectively the number of units and the number of iterations.
$O(MT)$ is the time complexity of GNG because $O(M)$ is required to find the best match unit of a presented data point at every iteration.
To decide the best match unit, we need to compute the distance between every reference vector and a presented data point and find the minimum distance.
Thus, we calculate and compare $M$ distances at every iteration.
$O(M^3)$ is the time complexity of SC to partition the units.
$O(MN)$ is the time complexity to assign data points to clusters.
To assign data points to clusters, we determine the nearest reference vector of every data point.
Thus, we calculate $M$ distances for each data point.
For $N \gg  M$, $O(MN)$ dominates the computational time of ASCs with GNG.

If we use a sort algorithm, such as quicksort, to simultaneously find the best match unit and the second-best match unit every iteration, the time complexity of ASC with GNG is $O(T M \log M + M^3 + NM)$.
In this case, GNG sorts the distances between all the reference vectors and a presented data point to finds the best match and the second-best match units at every iteration.
The sort algorithm takes $O(M \log M)$.

The time complexity of ASC with NG is $O(T M \log M + M^3 + NM)$.
The complexity of NG is $O(T M \log M)$.
In NG, the distances between all the reference vectors and a presented data point are calculated at every iteration.
The calculation of the distances takes $O(M)$.
NG sorts the distances to obtain the neighborhood ranking of the reference vectors at every iteration.
The sort of $M$ distances takes $O(M \log M)$ when we use quicksort.
NG updates the weights of all the reference vectors at every iteration.
NG takes $O(M)$ to update the weighs.
Thus, the time complexity of NG is $O(T(M + M \log M + M)) = O(T M \log M)$.
For $N \gg M$, $O(MN)$ dominates the computational time of ASC with NG.

The time complexity of ASC with SOM is $O(MT + M^3 + NM)$.
$O(MT)$ is the time complexities of SOM because it requires finding the best match unit of a presented data point and updating all the reference vectors at every iteration.
For $N \gg M$, $O(MN)$ dominates the computational time of ASCs with SOM.

The time complexities of ASC with \textit{k}-means is $O(MNT_\mathrm{kmeans} + M^3 + NM)$, where $T_\mathrm{kmeans}$ is the number of iterations in \textit{k}-means.
$O(MNT_\mathrm{kmeans})$ is the time complexities of \textit{k}-means because the distances between $M$ centroids and $N$ data points are calculated at every iteration.
It is difficult to estimate $T_\mathrm{kmeans}$, but the convergence of \text{k}-means is fast \citep{Bottou:1994} and $T_\mathrm{kmeans}$ will be enough smaller than $N$.
For $N \gg M$, $O(MNT_\mathrm{kmeans} + NM)$ dominates the computational time of ASCs with \textit{k}-means.
In this case, the computational time approximately linearly increases with $N$ because $T_\mathrm{kmeans}$ will be enough smaller than $N$.

The space complexity of SC is $O(N^2)$ \citep{Mall:2013} because the memory space is required to store the $N \times N$ similarity matrix.
The space complexity of ASCs with GNG, NG, SOM, and \textit{k}-means is $O(N + M^2)$.
$O(N)$ is the memory space to store the data points.
$O(M^2)$ is the memory space to store the similarity matrix of reference vectors.
For $N \gg M$, the space complexity of the ASCs is $O(N)$.
Thus, the space complexities of the ASCs are much smaller than that of SC.

The time complexities and the space complexities of ASCs and SC are summarized in Tab. \ref{tab:comp}.

\begin{table}
\begin{center}
  \caption{Time complexities and space complexities}
  \label{tab:comp}
\begin{tabular}{ |c|c|c| }
 \hline
 Method           & time complexity                     & space complexity \\ \hline
 ASC with GNG     & $O(MT + M^3 + NM)$                  & $O(N + M^2)$ \\ \hline
 ASC with NG      & $O(TM \log M + M^3 + NM)$           & $O(N + M^2)$ \\ \hline
 ASC     & $O(MT + M^3 + NM)$                  & $O(N + M^2)$ \\ \hline
 ASC with k-means & $O(MNT_\mathrm{kmeans} + M^3 + NM)$ & $O(N + M^2)$ \\ \hline
 SC               & $O(N^3)$                            & $O(N^2)$ \\ \hline
\end{tabular}
\end{center}
\end{table}

\subsection{Parameters}

All the parameters of the methods, except for $T$, $M_\mathrm{max}$, $M$, and $l$, are found using grid search to maximize the mean of the purity scores for the datasets used in Sec.\ref{sub:clustering_quality}.
The parameters of ASC with GNG are $T = 10^5$, $\lambda = 250$, $\varepsilon_1 = 0.1$, $\varepsilon_n = 0.01$, $a_\mathrm{max} = 75$, $\alpha = 0.25$, $\beta = 0.99$, $\sigma = 0.25$.
The parameters of ASC with NG are $T = 10^5$, $\lambda_i = 1.0$, $\lambda_f = 0.01$, $\varepsilon_i = 0.5$, $\varepsilon_f = 0.005$, $a_{\mathrm{max}i} = 100$, $a_{\mathrm{max}f} = 300$, and $\sigma = 0.25$.
The Parameters of ASC with SOM are $T = 10^5$, $\gamma_0 = 0.05$, $\sigma_0 = 1.0$, and $\sigma = 0.5$.
The Parameter of ASC with \textit{k}-means is $\sigma = 0.1$.
The parameters of ASC with GNG no edge are $T = 10^5$, $\lambda = 350$, $\varepsilon_1 = 0.05$, $\varepsilon_n = 0.01$, $a_\mathrm{max} = 100$, $\alpha = 0.5$, $\beta = 0.999$, $\sigma = 0.5$.
The Parameter of ASC with \textit{k}-means and SC is $\sigma = 0.1$.
The meanings of the parameters are shown in Sec.\ref{sub:algorithm} and Appendix.

\section{Results} % (fold)
\label{sec:results}

This section describes computational time and clustering performances of ASCs with GNG.

The programs used in the experiments were implemented using Python and its libraries.
The libraries are numpy for linear algebra computation, networkx for dealing a network, and scikit-learn for loading datasets and using \textit{k}-means.

\subsection{Computational time} % (fold)
\label{sub:computational time}

This subsection describes the computational time of ASCs with GNG, NG, SOM, and \textit{k}-means and SC.
For the measurement of the computational time, the dataset has five clusters and consists of 3-dimensional data points.
This dataset is generated by datasets.make\_blobs, which is the function of scikit-learn.
The dataset is called Blobs in this study.
In this experiment, I use the computer with two Xeon E5-2687W v4 CPUs and 64 GB of RAM and cluster the dataset using only one thread to not process in parallel.

\begin{figure*}[tb]
  \centering
    \includegraphics[width=140mm]{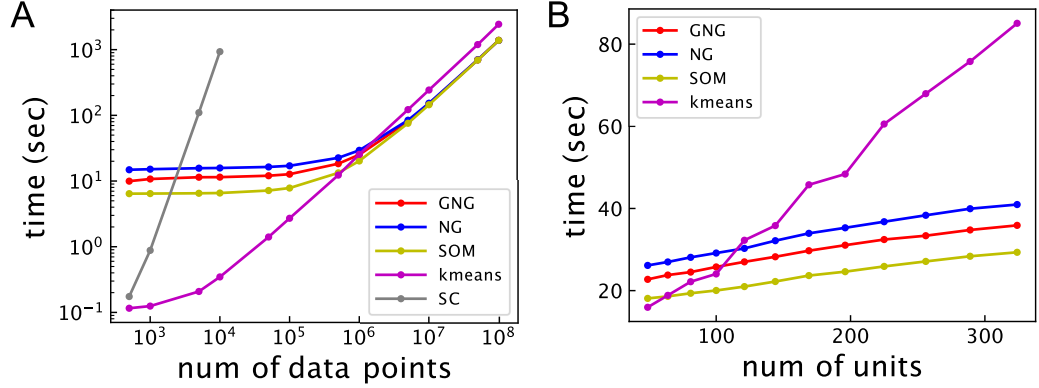}
    \caption{Computational time of ASCs for clustering the synthetic dataset. A. Relationship between computational time and the number of data points. The number of units of all ASCs is $10^2$. B. Relationship between computational time and the number of units. The number of the data points in the dataset is $10^6$.
GNG, NG, SOM, kmeans, and SC in these figures represent the computational time of ASCs with GNG, NG, SOM, and the \textit{k}-means and SC, respectively. The computational time is the mean of 10 runs with random initial values.
}
    \label{fig:computational_time}
\end{figure*}

Figure \ref{fig:computational_time}A shows the relationship between the computational time and the number of data points for $10^2$ units.
The computational time of ASC with \textit{k}-means is shortest under about $5 \times 10^5$ data points.
The computational time of ASC with \textit{k}-means linearly increases from about $10^4$ data points.
ASCs with GNG, NG, and SOM show better computational performance than ASC with \textit{k}-means from about $10^6$ data points.
The computational time of ASCs with GNG, NG, and SOM does not significantly change under about $10^5$ data points.
However, the computational time of ASCs with GNG, NG, and SOM linearly increases with the number of data points beyond about $10^6$ data points.
Under $10^5$ data points, the computational time of ASCs with GNG, NG, and SOM will mainly depend on the computational cost to generate a network because $O(MT)$ and $O(T M \log M)$ are more than $O(M^3)$ and $O(MN)$.
On the other hand, the linear increase of the computational time of ASCs with GNG, NG, and SOM will be dominantly derived from the computational cost to assign data points to units.
The computational time of ASC with SOM is shorter than that of ASCs with GNG and NG under about $10^6$ data points because SOM does not have the process to change the network topology.
From about $10^7$ data points, the computational time of ASC with SOM is not different from ASCs with GNG and NG because the computational cost to assign data points to units is dominant.
Under $10^6$ data points, the computational time of ASC with GNG is shorter than ASC with NG.
This difference occurs because the mean of growing $M$ of ASC with GNG during learning is less than $M$ of ASC with NG.
The computational time of SC is not shown for more than $10^4$ data points because the computational time of spectral clustering is too long compared to the other methods.
These results show that the ASCs outperform SC in terms of computational time for a large dataset.
However, the results also show that ASC using topology is not effective for a small dataset.

Figure \ref{fig:computational_time}B shows the relationship between the computational time and the number of units for $10^6$ data points.
In ASC with GNG, the number of units means the maximum of the number $M_\mathrm{max}$.
The computational time of all ASCs linearly increases with the number of units.
The computational time of the ASCs with GNG, NG, and SOM increases more slowly than \textit{k}-means because the complexity of \textit{k}-means depends on not only the number of units but also the number of data points.

% subsection computational time (end)

\subsection{Clustering quality} % (fold)
\label{sub:clustering_quality}

To investigate the clustering performances of the ASC with GNG, NG, and SOM, three synthetic datasets and six real-world datasets are used.
The synthetic datasets are Blobs, Circles, and Moons generated by datasets.make\_blobs, datasets.make\_circles, and datasets.make\_moons that are functions of scikit-learn, respectively.
Blobs are generated from three isotropic Gaussian distributions.
The standard deviation and the means of each Gaussian are default values of the generating function.
Blobs can be partition by a linear separation method such as \textit{k}-means.
Circles consists of two concentric circles.
The noise and the scale parameters of the function generating Circles are set at 0.05 and 0.5, respectively.
Moons includes two-moons shape distributed data points.
The noise parameter of the function generating Moons is 0.05.
Circles and Moons are typical synthetic datasets that can not be partitioned by a linear separation method.
The real-world data are Iris, Wine, Spam, CNAE-9, Digits, and MNIST \citep{Lecun:1998}.
Iris, Wine, Spam, CNAE-9, and Digits are datasets found in the UCI Machine Learning Repository.
In this study, Iris, Wine, Digits, and MNIST are obtained using scikit-learn.
Iris and Wine are datasets frequently used to evaluate the performance of a clustering method.
Spam and CNAE-9 are word datasets.
In this study, three attributions of Spam: capital\_run\_length\_average, capital\_run\_length\_longest, and capital\_run\_length\_total, are not used.
Digits and MNIST are handwritten digits datasets.
Table \ref{tab:datasets} shows the numbers of classes, data points, and attributions of the datasets.

\begin{table}[tb]
\caption{Datasets}
\begin{center}
\begin{tabular}{|c|c|c|c|}
\hline
Dataset &$k$ & $n$   & $d$           \\ \hline
Blobs   &  3 & 1000  & 2             \\ \hline
Circles &  2 & 1000  & 2             \\ \hline
Moons   &  2 & 1000  & 2             \\ \hline
Iris    &  3 & 150   & 4             \\ \hline
Wine    &  3 & 178   & 13            \\ \hline
Spam    &  2 & 4601  & 54            \\ \hline
CNAE-9  &  9 & 1080  & 856           \\ \hline
Digits  & 10 & 1797  & $8 \times 8$  \\ \hline
MNIST   & 10 & 70000 & $28 \times 28$\\ \hline
\end{tabular}
\begin{tablenotes}
$k$, $n$, and $d$ indicate the number of classes, data points and attributions, respectively.
\end{tablenotes}
\end{center}
\label{tab:datasets}
\end{table}

To evaluate the clustering methods, we use the purity score.
Purity is given by $\mathrm{Purity} = 1/N \sum^k_{i = 1} \max_{j} n^j_i$, where $N$ is the number of data points in a dataset, $k$ is the number of clusters, and $n^j_i$ is the number of data points that belong to the class $j$ in the cluster $i$.
When the purity score is 1, all data points belong to true clusters.

Table \ref{tab:purity} shows the purities of ASCs with GNG, NG, SOM, \textit{k}-means, and GNG using no topology and SC.
ASC with GNG shows the best accurate clustering results for five datasets: Circles, Moons, Spam, CNAE-9, and Digits.
For Blobs, ASC with GNG displays the second-best performance, but the difference of purity between ASC with GNG, NG, and GNG using no topology and SC is small.
For MNIST, ASC with GNG shows the second-best performance.
ASC with NG also shows relatively high purities for Blobs, Circles, Moons, and Digits.
However, for Iris and Wine dataset, the clustering performances of ASCs with GNG and NG are worse than the others.
The lower performance of ASCs with GNG and NG may be caused by too many units for the number of data points.
Perhaps, there will be the optimal number of units to make a better similarity matrix.
The performance of ASC with SOM is worse than the other methods for the datasets apart from Spam.
The bad performance of ASC with SOM may be caused by the feature of SOM that is the tendency to have null units often.
This feature of SOM is unsuitable for ASC using topology.
ASC with \textit{k}-means shows relatively high performance for Blobs, Circles, Moons, Iris, Wine, and MNIST.
For MNIST, the performance of ASC with \textit{k}-means is best.
ASC with GNG using no topology shows the best performance for Blobs, Circles, and Wine.
For Moons and Spam, ASC with GNG using no topology displays the second-best performance.
For Blobs, Circles, Moons, Wine, Spam, CNAE-9, and Digits, the performance of ASC with GNG using no topology is better than that of ASC with \textit{k}-means.
This result suggests that GNG can quantize dataset better than \textit{k}-means in many cases.
For MNIST, SC cannot perform clustering because overflow occurs.
These results suggest that the network topology effectively improves the performance of the clustering, and GNG will generate the same or better quantization result than \textit{k}-means.

\begin{table*}[tb]
\caption{Performances of clustering (purity)}
\begin{center}
\begin{tabular}{|c|c|c|c|c|c|c|c|c|c|c|c|}
\hline
dataset & GNG          & NG          & SOM   & \textit{k}-means & GNG               &  SC       \\
        &              &             &       &                  & using no topology &           \\ \hline
Blobs   & 0.9744       &0.9632       &0.4110 &0.9690            &{\bf 0.9813}        &0.9671 \\ \hline
Circles & {\bf 1.0000} &{\bf 1.0000} &0.5403 &0.9997            &{\bf 1.0000}       &{\bf 1.0000} \\ \hline
Moons   & {\bf 0.9992} &0.9884       &0.5810 &0.9328            &0.9985             &0.9933 \\ \hline
Iris    & 0.5840       &0.5648       &0.5020 &0.8473            &0.8427             &{\bf 0.8533} \\ \hline
Wine    & 0.4650       &0.4649       &0.4379 &0.6656            &{\bf 0.6827}       &0.6742 \\ \hline
Spam    & {\bf 0.7676} &0.6063       &0.6082 &0.6095            &0.7464             &0.6070 \\ \hline
CNAE-9  & {\bf 0.6711} &0.5920       &0.2913 &0.4887            &0.5706             &0.1871 \\ \hline
Digits  & {\bf 0.8572} &0.8129       &0.3356 &0.7641            &0.8025             &0.6023 \\ \hline
MNIST   & 0.6100       &0.5801       &0.2784 &{\bf 0.6754}      &0.5888             &nan \\ \hline
\end{tabular}
\begin{tablenotes}
GNG, NG, SOM, k-means, GNG using no topology, and SC mean ACSs with GNG, NG, SOM, k-means, and GNG using no topology, and spectral clustering, respectively.
The purities are the mean of 100 runs with random initial values. The best estimations are bold.
\end{tablenotes}
\end{center}
\label{tab:purity}
\end{table*}

\begin{figure*}[tb]
  \centering
    \includegraphics[width=140mm]{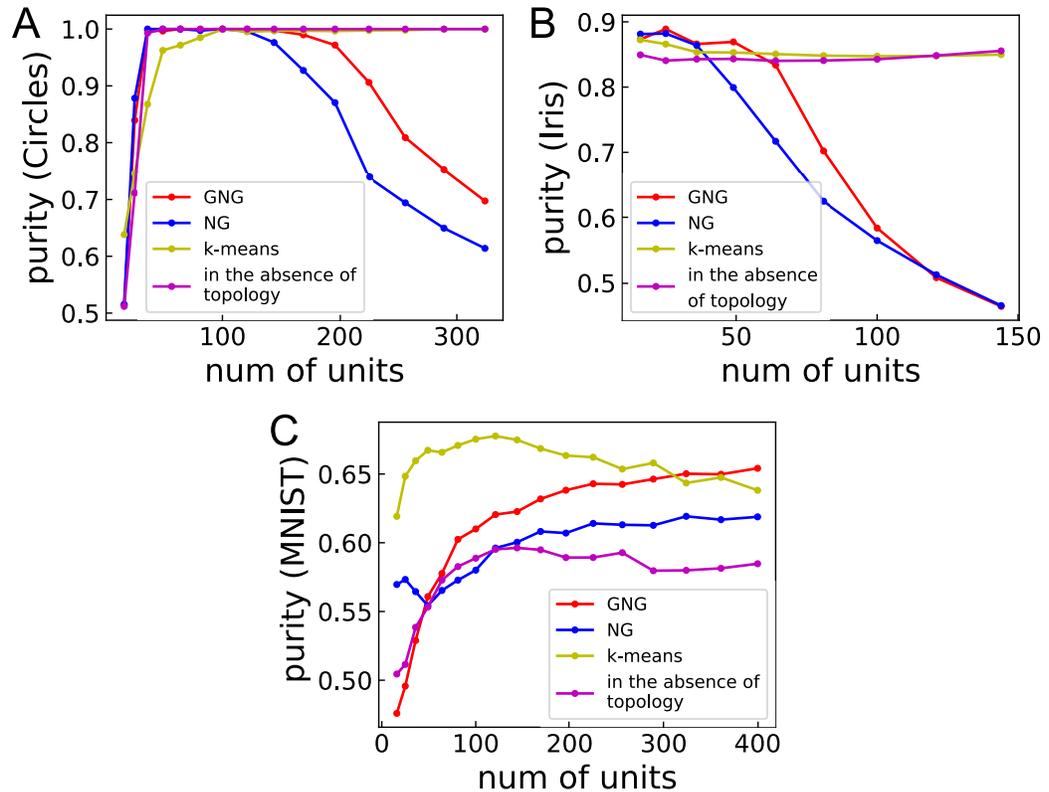}
    \caption{The relationship between purity and the number of units. A, B, and C show the performance of clustering Circles, Iris, and MNIST, respectively. Purities are the mean of 100 runs with random initial values. The line labeled ``in the absence of topology'' denotes the clustering performance of ASC with GNG using no topology.
    }
    \label{fig:purity_vs_units}
\end{figure*}

Figure \ref{fig:purity_vs_units} shows the relationship between purity and the number of units for Circles, Iris, and MNIST to investigate the dependence of clustering performance on the number of units.
For Circles, the clustering performances of ASCs with GNG and NG vary in convex upward.
This result suggests that the performances become worse for the larger or smaller number of units than the optimal number of units and become better for the around optimal number of units.
For Iris, the performances of ASCs with GNG and NG are high under about 50 and 36 units, respectively.
From about 50 and 36 units, these performances go down with the number of units.
This result suggests that the number of units used in Tab.  \ref{tab:purity} is too large to obtain the optimal performances for Iris and Wine.
For MNIST, the performances of ASCs with GNG and NG improve with the number of units.
For all cases, the performances of ASCs with \textit{k}-means and GNG using no topology more weakly depend on the number of units.
This result suggests that ASC not using network topology displays peak performance less than ASCs using network topology in many cases but its performance does not strongly depend on the number of units.

\section{Conclusion} % (fold)
\label{sec:conclusion}

This study proposes approximate spectral clustering using the network generated by growing neural gas, called ASC with GNG.
ASC with GNG partitions a dataset using a Laplacian matrix calculated from not only the reference vectors but also the topology of the network generated by GNG.
ASC with GNG displays better computational performance than SC for a large dataset.
Furthermore, the clustering quality of ASC with GNG is equal or better than SC in many cases.
The results of this study suggest that ASC with GNG improves not only computational but also clustering performances.
Therefore, ASC with GNG can be a successful method for a large dataset.

Why does ASC with GNG display better clustering performance?
The clustering results of SC depend on the quality of the constructed network from which a Laplacian matrix is calculated \citep{Li:2018}.
Namely, we need to improve the way to construct a similarity matrix to obtain better clustering performance for SC \citep{Lu:2016,Park:2018}.
The network generated by GNG represents the important topological relationships in a given dataset \citep{Fritzke:1995}.
Furthermore, in ASC with GNG, a similarity matrix calculated using the network topology is sparse because the elements of a similarity matrix between not connected units are zero.
The sparse similarity matrix may contribute to the improvement of clustering performance.
Therefore, the ability of GNG to extract the topology of a dataset will lead to improving clustering performance.

However, when the number of units is not optimal, ASC with GNG will produce more unsatisfactory results than ASC with GNG using no topology, and SC.
This problem occurs when the number of units is close to or too smaller than the number of data points.
This problem will not frequently occur in the actual application dealing with a large dataset.

ASC using no topology, such as ASCs with k-means and GNG using no topology, also displays high performance, but its peak performance is less than ASC with GNG in many cases.
If we are concerned about the dependence of performance on the number of units, then ASC using no topology may also be a good option.

Why can ASC partition a nonlinearly separable dataset?
The ability for nonlinear clustering is provided by SC used in abstraction level 2.
Simply put, abstraction level 1 reduces the size of a dataset.
The assignment after abstraction level 2 does not provide nonlinear separation, as only it creates only Voronoi regions around the reference vectors.
SC makes the complex decision region merging the Voronoi regions and achieves nonlinear clustering.
Thus, the ability for nonlinear clustering is not derived from the way of abstraction level 1.
However, the clustering performance of ASC also depends on the way of abstraction level 1, as mentioned in the results section.

ASC with GNG can be more accelerated by parallel computing.
\cite{Garcia-RodriGuez:2011} have achieved to accelerate GNG using graphics processing unit (GPU).
\cite{Vojacek:2013} have parallelized GNG algorithm using high-performance computing.
Thus, the process of making the network using GNG can be accelerated by GPU or high-performance computing.
Furthermore, finding the nearest reference vectors of data points can easily be parallelized because each calculation of distance is independent.
If the nearest reference vectors are found using parallel computing with $p$ threads, the computational time of the finding is reduced to $1/p$.

SC and ASC have a limitation as well as other clustering methods such as k-means.
SC and ASCs can make nonlinear boundaries but can only group neighboring data points based on similarity or distance.
In other words, they cannot group some not-neighboring subclusters that represent the same category.
For instance, in MNIST, the subclusters representing the same digit are scattered, as shown in Fig. 10 of \cite{Khacef:2019}.
To accurately cluster such a dataset, the scattered subclusters must be merged into one cluster.
Khacef et al. \citep{Khacef:2019,Khacef:2020} have accurately clustered MNIST using only six hundred labeled data points.
Their proposed method projects input data onto a 2-dimensional feature map using SOM and then labels reference vectors using post-labeled unsupervised learning.
Post-labeled unsupervised learning using SOM is efficient if we have some labeled data points in a dataset and require a more accurate clustering result.

\appendix
\label{sec:appendix}

\section{Self-organizing map and its alternatives}

Self-organizing map (SOM) and its alternatives, such as NG and GNG, are artificial neural networks using unsupervised learning.
They convert data points to fewer weights (representative vectors) of units in a neural network and preserve the input topology as the topology of the neural network.
Kohonen's SOM \citep{Kohonen:1990} is a basic and typical SOM algorithm.
Kohonen's SOM has the features that the network topology is fixed into a lattice \citep{Sun:2017} and the number of units is constant.
NG has been proposed by Martinetz and Schulten \citep{Martinetz:1991} and can flexibly change the topology of its network.
However, in NG, we have to preset the number of units in the network.
GNG \citep{Fritzke:1995} achieves to flexibly change both the network topology and the number of units in the network according to an input dataset.
GNG can find the topology of an input distribution \citep{Garcia-RodriGuez:2012}.
GNG has been widely applied to clustering or topology learning such as extraction of two-dimensional outline of an image \citep{Angelopoulou:2011,Angelopoulou:2018}, reconstruction of 3D models \citep{Holdstein:2008}, landmark extraction \citep{Fatemizadeh:2003}, object tracking \citep{FrezzaBuet:2008}, and anomaly detection \citep{Sun:2017}.

\subsection{Kohonen's Self-organizing map}

Kohonen's SOM is one of the neural network algorithms and a competitive learning and unsupervised learning algorithm.
Kohonen's SOM can project high-dimensional to a low-dimensional feature map.
The function of the projection is used for cluster analysis, visualization, and dimension reduction of a dataset.

The topology of the network of Kohonen's SOM is a two-dimensional $l \times l$ lattice in this study, where $l \times l = M$.
The unit $i$ in the network has the reference vector $\bm{w}_i \in R^d$.
The unit $i$ is at $\bm{p}_i \in R^2$ on lattice, where $\bm{p}_i = ((\mathrm{mod}(i - 1, l) + 1)/l, \lceil i/l \rceil/l), i = \{1, 2, ..., M\}$.
$\mathrm{mod}(a, b)$ is the remainder of the division of $a$ by $b$.
A general description of SOM algorithm is as follows:
\begin{enumerate}
 \item Initialize the reference vectors of the units.
 All elements of the reference vectors are randomly initialized in the range of [0, 1].
 \item Randomly select a data point $\bm{x}_n$ and find the best match unit $c$, that is
 \begin{equation}
 c = \argmin_i \| \bm{x}_n - \bm{w}_i \|.
 \end{equation}
 \item Update the reference vectors of all units. The new reference vector of the unit $i$ is defined by
 \begin{equation}
 \bm{w}_i \leftarrow \bm{w}_i + h_{ci}(t)(\bm{x}_n - \bm{w}_i),
 \end{equation}
 where $t$ is the number of iterations, $h_{ci}(t)$ is the neighborhood function.
 $h_{ci}(t)$ is described by the following equation:
 \begin{equation}
 h_{ci}(t) = \alpha(\gamma_0, t) \exp \Big (- \frac{\mathrm{sqdist}(i, c)}{2 \alpha(\sigma_0, t)^2} \Big),
 \end{equation}
 where $\alpha(z, t)$ is a monotonically decreasing scalar function of $t$, and sqdist$(i, c)$ is the square of the geometric distance between the unit $i$ and the best match unit $c$ on the lattice.
 $\alpha(z, t)$ is defined by
 \begin{equation}
 \alpha(z, t) = z \times (1 - \frac{t}{T}),
 \end{equation}
 $\mathrm{sqdist}(i, c)$ is defined by
 \begin{equation}
 \mathrm{sqdist}(i, c) = \|\bm{p}_i - \bm{p}_c\|^2.
 \end{equation}
 \item If $t = T$, terminate. Otherwise, go to Step 2.
\end{enumerate}

\subsection{Neural gas (NG)} % (fold)
\label{sub:neural_gas}

NG also generates a network from input data points.
NG flexibly changes the topology of the network according to input data points, but the number of the units is static.
The network consists of $M$ units and edges connecting pairs of units.
The unit $i$ has the reference vector $\bm w_i$.
The edges are not weighted and not directed.
An edge has a variable called age to decide whether the edge is deleted.
Let us consider the set of $N$ data points, $X = \{\bm x_1, \bm x_2, ..., \bm x_n,..., \bm x_{N}\}$, where $\bm x_n \in R^d$.
The algorithm of the neural gas is shown below:
\begin{enumerate}
 \item Assign initial values to the weight $\bm{w}_i \in R^d$ and set all $C_{ij}$ to zero.
 $C_{ij}$ describes the connection between the unit $i$ and the unit $j$.
 \item Select a data point $\bm{x}_n$ from the dataset at random.
 \item Determine the neighborhood-ranking of $i$, $k_i$, according to distance between $\bm{w}_i$ and $\bm{x}_n$ by the sequence of ranking $(i_0, i_1, ..., i_k, ..., i_{M-1})$ of units with
 \begin{equation}
 \| \bm{x} - \bm{w}_{i_0} \| < \| \bm{x} - \bm{w}_{i_1} \| < ... < \| \bm{x} - \bm{w}_{i_k} \| < ... < \| \bm{x} - \bm{w}_{i_{N-1}} \|.
 \end{equation}
 \item Perform as adaptation step for the weights according to
 \begin{equation}
 \bm{w}_i \leftarrow \bm{w}_i + \varepsilon e^{- k_i/\lambda}(\bm{x}_n - \bm{w}_i), i = 1,...,N.
 \end{equation}
 \item Determine the nearest neighbor unit $i_0$ and the second nearest neighbor unit $i_1$.
 If $C_{i_0i_1}$ = 0, set $C_{i_0i_1} = 1$ and $l_{i_0i_1} = 0$.
 If $C_{i_0i_1} = 1$, set $l_{i_0i_1} = 0$.
 $l_{i_0i_1}$ describes the age of the edge between the unit $i_0$ and the unit $i_1$.
 \item Increase the age of all connections of $i_0$ by setting $l_{i_0j} = l_{i_0j} + 1$ for all $j$ with $C_{i_0j}$ = 1.
 \item Remove all connections of $i_0$ that have their age exceeding the lifetime by setting $C_{i_0j} = 0$ for all $j$ with $C_{i_0j} = 1$ and $l_{i_0j} > a_\mathrm{max}$.
 \item If the number of iterations is not $T$, go to step 2.
\end{enumerate}
$\varepsilon$, $\lambda$, and $a_\mathrm{max}$ decay with the number of iterations $t$.
This time dependence has the same form for these parameters and is determined by $g(t) = g_i (g_f/g_i)^{t/T}$.

% subsection neural_gas (end)

\subsection{Growing neural gas}
\label{sec:gng}

Growing neural gas (GNG) can generate a network from a given set of input data points.
The network represents important topological relations in the data points using Hebb-like learning rule \citep{Fritzke:1995}.
A network generated by GNG consists of units and edges that are connections between units.
In GNG, not only the number of the units but also the topology of the network can flexibly change according to input data points.
The unit $i$ has the reference vector $\bm w_i \in R^d$ and summed error $E_i$.
The edges are not weighted and not directed.
An edge has a variable called age to decide whether the edge is deleted.

Let us consider the set of $N$ data points, $X = \{\bm x_1, \bm x_2, ..., \bm x_n,..., \bm x_{N}\}$, where $\bm x_n \in R^d$.
The algorithm of the GNG to make a network from $X$ is given by the following:
\begin{enumerate}
 \item Start the network with only two units that are connected to each other.
 The reference vectors of the units set two data points randomly selected from $X$.
 \item Select a data point $\bm{x}_n$ from the dataset at random.
 \item Find the winning unit $s_1$ of $\bm{x}_n$ by
 \begin{equation}
 s_1 = \arg \min_i \| \bm{x}_n - \bm{w}_i\|.
 \end{equation}
 Simultaneously, find the second nearest unit $s_2$.
 \item Add the squared distance between $\bm{x}_n$ and $\bm{w}_{s_1}$ to the summed error $E_{s_1}$:
 \begin{equation}
 E_{s_1} \leftarrow E_{s_1} + \|\bm{x}_n - \bm{w}_{s_1}\|.
 \end{equation}
 \item Move $\bm{w}_{s_1}$ toward $\bm{x}_n$ by fraction $\varepsilon_{s_1}$ of the total distance:
 \begin{equation}
 \bm{w}_{s_1} \leftarrow \bm{w}_{s_1} + \varepsilon_{s_1} (\bm{x}_n - \bm{w}_{s_1}).
 \end{equation}
 Also move the reference vectors of the all direct neighbor units $s_n$ of $s_1$ toward $\bm{x}_n$ by the fraction $\varepsilon_{s_n}$ of the total distance:
 \begin{equation}
 \bm{w}_{s_n} \leftarrow \bm{w}_{s_n} + \varepsilon_{s_n} (\bm{x}_n - \bm{w}_{s_n}).
 \end{equation}
 \item If $s_1$ and $s_2$ are connected by an edge, set the age of this edge to zero. 
 If $s_1$ and $s_2$ are not connected, create the edge connecting between these units.
 \item Add one to the ages of all the edges emanating from $s_1$.
 \item Remove the edges with their age larger than $a_\mathrm{max}$. If this results in nodes having no emanating edges, remove them as well.
 \item Every certain number $\lambda$ of the input data point generated, insert a new unit as follows:
 \begin{itemize}
 \item Determine the unit $q$ with the maximum summed error $E_q$.
 %\item Find the farthest node $f$ among the neighbors of $q$.
 \item Find the node $f$ with the largest error among the neighbors of $q$.
 \item Insert a new unit $r$ halfway between $q$ and $f$ as follows:
 \begin{equation}
 \bm{w}_r = (\bm{w}_q + \bm{w}_f)/2.
 \end{equation}
 The number of units has the limit $M_\mathrm{max}$.
 \item Insert edges between $r$ and $q$, and $r$ and $f$. Remove the edge between $q$ and $f$.
 \item Decrease the summed errors of $q$ and $f$ by multiplying them with a constant $\alpha$.
 Initialize the summed error of $r$ with the new summed error of $q$.
 \end{itemize}
 \item Decrease all summed errors by multiplying them with a constant $\beta$
 \item If the number of iterations is not $T$, go to step 2.
\end{enumerate}

%\bibliographystyle{spbasic}
%\bibliography{clustering,som,books,highdimension}

\end{document}